\documentclass[11pt]{article}

\usepackage[preprint]{acl}

\usepackage{times}
\usepackage{latexsym}
\usepackage{amsmath}

\usepackage{listings}
\usepackage{xcolor} % 如果你想要彩色语法高亮

\usepackage{tcolorbox}
\tcbset{
    mypromptstyle/.style={
        colback=gray!10,       % 背景颜色
        colframe=gray!80,      % 边框颜色
        fontupper=\small,      % 字体大小
        boxsep=5pt,            % 内边距
        left=5pt,
        right=5pt,
        top=5pt,
        bottom=5pt,
        breakable,             % 自动换行并可分页
        enhanced,              % 支持更多功能
    }
}

\usepackage[T1]{fontenc}
\usepackage[utf8]{inputenc}

\usepackage{microtype}

\usepackage{inconsolata}

\usepackage{graphicx}

\title{CoAuthorAI: A Human in the Loop System For Scientific Book Writing}

\author{
\begin{tabular}{c}
\textbf{Yangjie Tian}$^{1,2}$ \quad
\textbf{Xungang Gu}$^{1,3}$ \quad
\textbf{Yun Zhao}$^{1}$ \quad
\textbf{Jiale Yang}$^{1}$ \\
\textbf{Lin Yang}$^{1}$ \quad
\textbf{Ning Li}$^{1}$ \quad
\textbf{He Zhang}$^{1}$\thanks{Corresponding author.} \quad
\textbf{Ruohua Xu}$^{1}$ \\
\textbf{Hua Wang}$^{2}$ \quad
\textbf{Kewen Liao}$^{3}$ \quad
\textbf{Ming Liu}$^{3}$\footnotemark[1]
\end{tabular}
\\[6pt]
$^{1}$ Kexin Technology, Beijing 100012, China \\
$^{2}$ Institute for Sustainable Industries and Liveable Cities, Victoria University, VIC 3011, Australia \\
$^{3}$ School of Information Technology, Deakin University, Melbourne, VIC 3125, Australia \\
\texttt{yangjie.tian@live.vu.edu.au, zhanghe@kxsz.net, m.liu@deakin.edu.au}
}

\begin{document}
\maketitle
\begin{abstract}
Large language models (LLMs) are increasingly used in scientific writing but struggle with book-length tasks, often producing inconsistent structure and unreliable citations. We introduce \emph{CoAuthorAI}, a human-in-the-loop writing system that combines retrieval-augmented generation, expert-designed hierarchical outlines, and automatic reference linking. The system allows experts to iteratively refine text at the sentence level, ensuring coherence and accuracy. In evaluations of 500 multi-domain literature review chapters, \emph{CoAuthorAI} achieved a maximum soft-heading recall of 98\%; in a human evaluation of 100 articles, the generated content reached a satisfaction rate of 82\%. The book \textit{AI for Rock Dynamics}  generated with CoAuthorAI and Kexin Technology’s LUFFA AI model has been published with Springer Nature. These results show that systematic human–AI collaboration can extend LLMs’ capabilities from articles to full-length books, enabling faster and more reliable scientific publishing. The system demonstration video is available at \url{https://youtu.be/PAWQz48tsdA}.

\end{abstract} 

\section{Introduction}
Scientific writing is essential but complex and time-consuming. Writing books requires extensive research, careful organization, and multiple revisions. Large language models (LLMs) offer new ways to speed up and improve this process, from generating drafts to suggesting edits, saving authors significant time.

Recent work shows that LLMs such as ChatGPT, GPT-4, and Claude have begun to reshape short-form scientific text generation, including literature summarization \citep{agarwal2024litllm,wang2024autosurvey}, report drafting \citep{aljamaan2024reference,taylor2022galactica,r2gengpt2023}, and even chapter writing \citep{schoenenberger2023gptbook}. In these scenarios, fluent prose is produced rapidly while retrieval modules provide up-to-date facts, yielding tangible productivity gains. Nevertheless, fully automatic generation frequently suffers from citation hallucinations, factual inaccuracies, and stylistic inconsistencies across long documents \citep{alkaissi2023hallucination}.

To mitigate these shortcomings, the research community has embraced the human-in-the-loop (HITL) paradigm, where human expertise is interleaved with model inference for planning, content vetting, and approval \citep{hsu2024chime,agarwal2024litllm}. This paradigm leverages the complementary strengths of humans (domain knowledge, critical judgment) and LLMs (linguistic fluency, rapid drafting), and is becoming the de facto standard for high-stakes scientific communication.

Despite notable advances in short-form writing, the systematic development of long-form book drafting remains substantially underexplored. Existing prototypes such as BetaWriter’s fully automated monograph \citep{springer2019betawriter} and Meta’s Galactica demonstration \citep{taylor2022galactica} illustrate both the potential benefits and the significant limitations of generating book length manuscripts without sustained expert oversight. In practice, large publishers still rely on labor-intensive editing cycles to maintain coherence, control narrative depth, and ensure traceable citations \citep{schoenenberger2023gptbook}. This raises a key question: how can we scale LLM assistance to book-length writing while keeping human authors firmly in control?

To address this challenge, we present CoAuthorAI, a production-ready HITL system for scientific book generation. Our contributions are threefold:

\begin{enumerate}
    \item Design a \emph{modular architecture} combining retrieval-augmented generation, expert-designed hierarchical outlines, and automatic reference linking, enabling chapter-level generation with sentence-level traceability.
    \item Implement \emph{interactive feedback loops} that let experts iteratively refine outlines, regenerate sections, and verify citations, ensuring control over style, depth, and accuracy.
    \item Explore the boundaries between LLMs and domain experts in book-writing tasks, and using this system in collaboration with Kexin Technology’s LUFFA model to assist author teams in publishing \textit{AI for Rock Dynamics} \footnote{https://link.springer.com/book/10.1007/978-981-96-5342-3?sap-outbound-id=D94D3E307CE1F96013B03FB247B741415100E16B}.

\end{enumerate}

Collectively, these advances extend collaborative writing with LLMs from articles to full-length books, providing a practical workflow for authors and publishers.

\section{Related Work}
We organise the discussion around three strands of research that converge on human‑in‑the‑loop book generation.

\paragraph{Literature Summarisation}
Early systems like \textit{LitLLM} \citep{agarwal2024litllm} and \textit{AutoSurvey} \citep{wang2024autosurvey} employ retrieval‑augmented generation pipelines to convert collections of papers into structured literature reviews. Interactive platforms such as \textit{Elicit} \citep{elicit2024} and \textit{SciSpace} \citep{scispace2024} extend this idea, offering query‑driven paper discovery and summary previews that researchers can curate manually. These works demonstrate the efficiency gains of combining search and generation but are typically limited to section‑scale outputs.

\paragraph{Scientific Report Generation}
General‑purpose models (e.g.\ GPT‑3.5/4, Claude) have been tested for writing grant proposals and clinical reports, yet high hallucination rates in references remain a major challenge \citep{alkaissi2023hallucination}. Domain‑constrained approaches leverage multimodal grounding (e.g.\ \textit{R2GenGPT} for radiology) or curated corpora (Elsevier’s \textit{ScienceDirect AI}) to improve factuality \citep{r2gengpt2023,elsevier2025sciencedirect}. While these systems showcase HITL verification interfaces, they tackle documents far shorter than a full book.

\paragraph{Book Drafting and HITL Workflows}
 Beta Writer initiated the study of automatic book construction by clustering and summarising existing academic articles, although the resulting prose exhibited substantial fragmentation and required extensive human post editing\citep{springer2019betawriter}. Subsequent industrial efforts have explored the integration of conversational large language models into collaborative authoring pipelines. For example, Springer Nature’s GPT assisted textbook reportedly reduced production time by 50\% while retaining human authorship as the central decision making component \citep{schoenenberger2023gptbook}. However, publicly available technical accounts detailing system architectures, evaluation protocols and design trade offs remain limited, leaving key questions unresolved concerning scalability, citation reliability and the granularity of expert involvement. Our work addresses this gap by providing a fully documented system together with a comprehensive empirical analysis.

\begin{figure}
    \centering
    \includegraphics[width=1\linewidth]{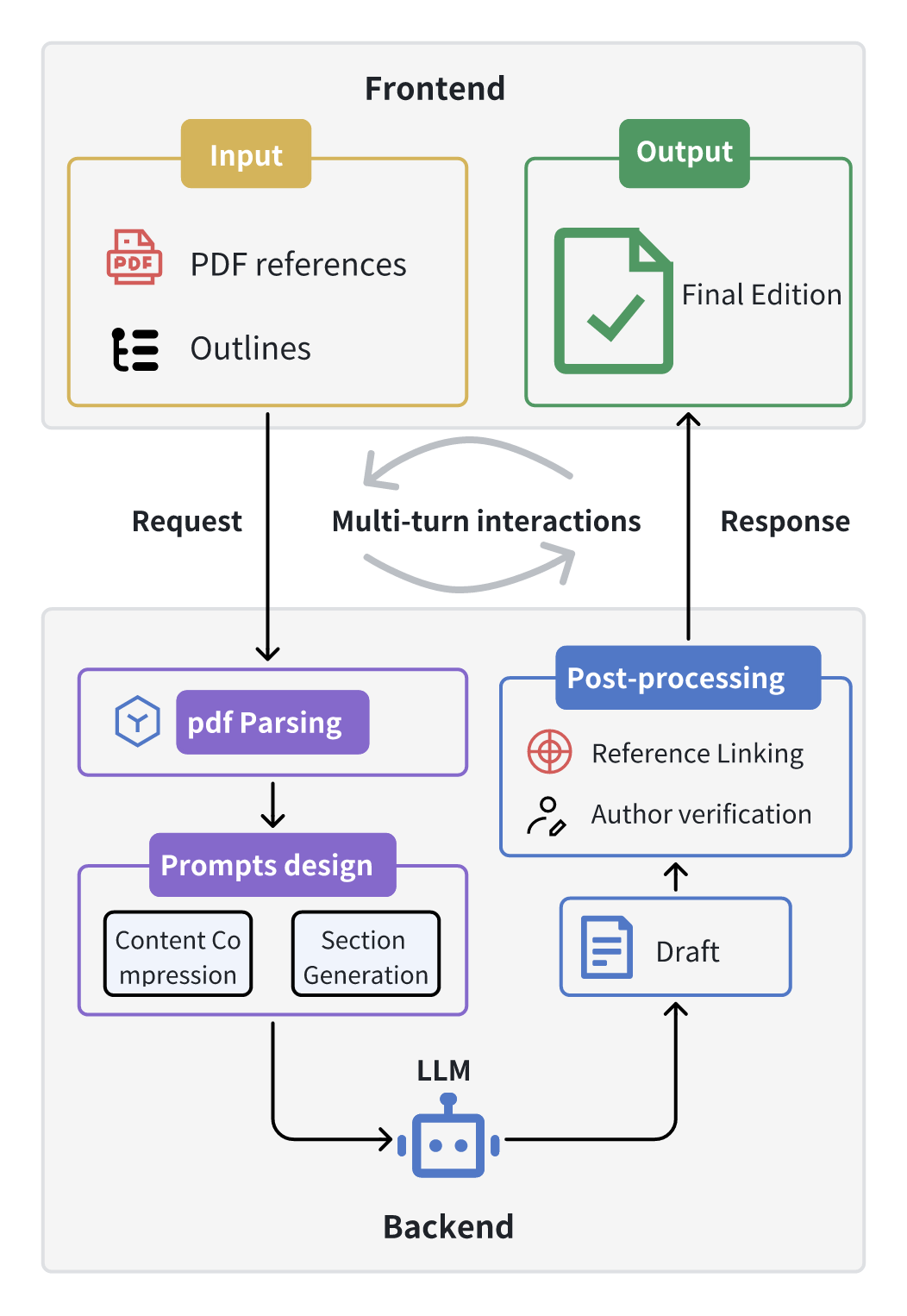}
    \caption{Overview of CoAuthorAI, illustrating the frontend for expert inputs and revision, and the backend for PDF parsing, content compression, section generation, and post-processing through LLM interaction.}
    \label{fig:overview}
\end{figure}

\section{CoAuthorAI}

Our CoAuthorAI demonstration system is a web application built with Streamlit\footnote{https://streamlit.io/} and employs Python\footnote{https://www.python.org/} for preprocessing tasks. We leverage the PDF parsing tool for extracting content from PDF documents, and store the resulting embeddings in Milvus\footnote{https://milvus.io/zh}. The system is composed of two primary components: a front-end and a back-end. As depicted in Figure~\ref{fig:overview}, The front-end handles user interactions, such as document uploads, book outline creation, and expert content revisions. On the back-end, we use the PDF parsing tool to convert PDF articles into machine-readable formats and, combined with retrieval-augmented techniques, employ large language models to generate chapter content. Furthermore, we provide a detailed, step-by-step guide to using CoAuthorAI.

\begin{figure}[ht]
    \centering
    \includegraphics[width=0.49\textwidth]{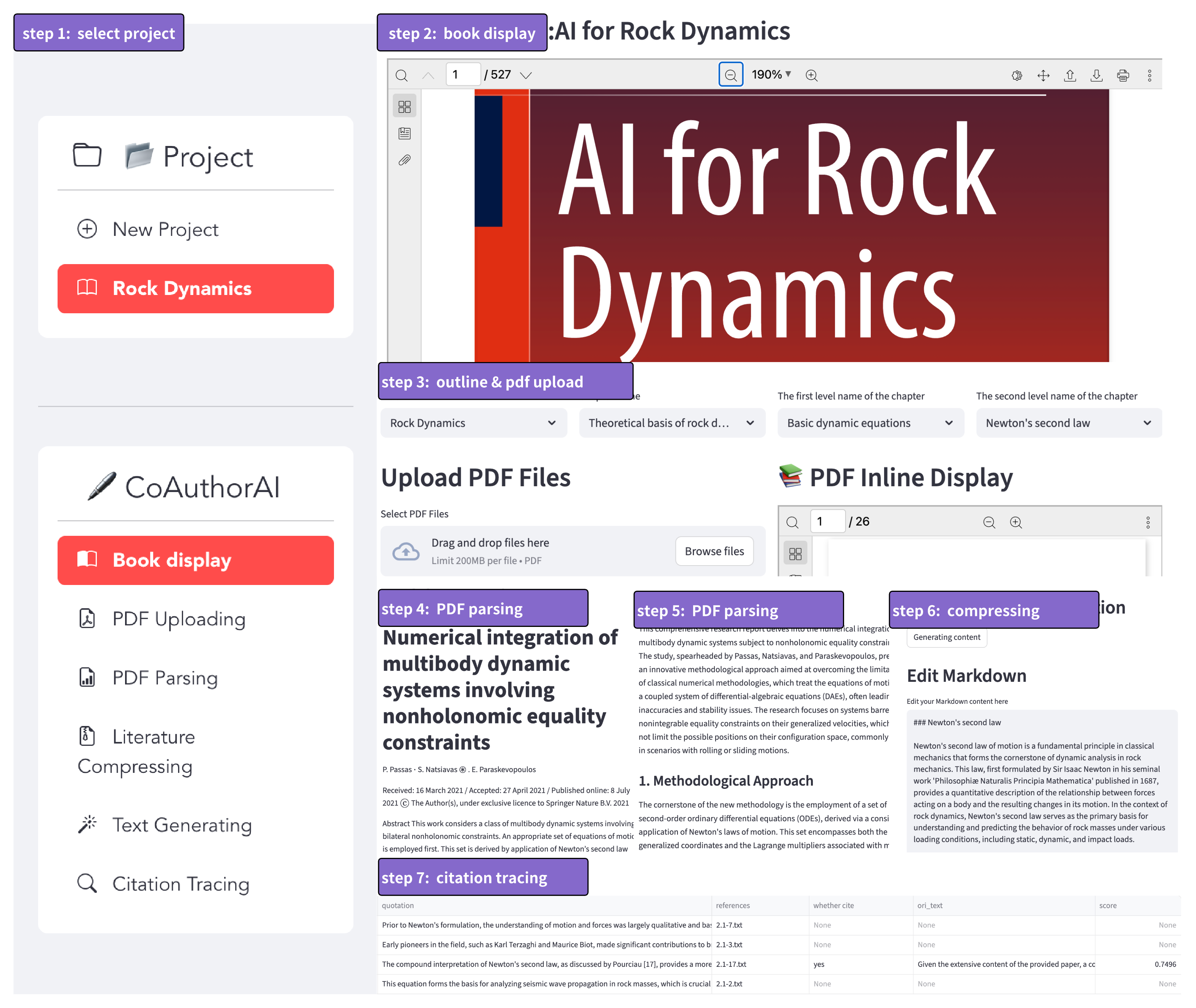}
    \caption{User interface and workflow of the CoAuthorAI system. The interface guides users through a seven-step human–AI collaborative pipeline: (1) selecting or creating a project, (2) viewing and managing book-level information, (3) uploading outlines and reference PDFs, (4) parsing PDF content , (5) compressing literature, (6) manual refinement of AI-generated text, and (7) performing citation tracing.}
    \label{fig:demonstration}
\end{figure}

\subsection{Frontend Design}
\paragraph{Human Guidance and Looping}Human guidance and iterative feedback are essential for ensuring the quality and accuracy of AI-generated content. As shown in Figure~\ref{fig:demonstration}, experts are responsible for establishing the book’s title and outline, and upload relevant literature for each chapter to provide direction and factual material. Because LLMs outputs can be random and machine-like, experts review, refine, and verify the generated content on the platform to meet academic and industry standards. This involves multiple rounds of human intervention and feedback loops to continuously improve the AI-generated text.

\subsection{Backend Design}
\paragraph{PDF Extraction} Using a self-developed PDF parsing tool, which is designed to efficiently analyze and extract content from PDF documents that contain complex elements such as images, formulas, and tables. It converts multi-modal PDFs into Markdown format, making them easier to process and suitable for computational analysis and machine learning tasks. With the help of this tool, users can construct large language model (LLM) corpora more effectively and make full use of the complete information contained in PDF materials.
\paragraph{Content Compression}In experiments with multi-document long-form content generation, we found two major limitations of LLMs. First, the context window acts as the model’s working memory, limiting the size of documents it can process at once, which is often smaller than the total reference material for a book. Second, even with techniques to expand the context, generated content tends to focus on conceptual explanations and lacks detailed coverage of specific knowledge. To address this, we first use LLMs to compress and summarize full-text documents, then feed the processed text as reference. This improves generation efficiency, depth, and accuracy.

\paragraph{Section Generation}The section generation module serves as the core of our content production pipeline. We employ LLMs to generate section content based on compressed reference materials. We observed that more detailed prompts lead to higher-quality outputs, so domain experts refine content outlines down to second-level subheadings before generation. Due to context window limitations and after repeated experiments, we found that to ensure in-depth discussion, the number of compressed reference materials should not exceed 40 documents. For sections requiring more references, our backend system uses a multi-stage generation approach:

\begin{itemize}
    \item \textbf{Step 1:} Using the same prompt engineering, we generate the content of the chapter for every 40 reference documents, treating each as an intermediate result;
    \item \textbf{Step 2:} After batch generation, these intermediate results are used as reference materials to produce the final version of the chapter;
    \item \textbf{Step 3:} The final generated chapter is then linked back to the original materials to establish correspondence.
\end{itemize}
This batch-processing architecture ensures maximum utilization of reference materials while maintaining coherent narrative flow.

\paragraph{Reference Linking} module plays a crucial role in ensuring the accuracy and credibility of citations in the generated content. To achieve this, we first decompose the source documents into smaller semantic blocks, and then transform both the generated content and the references into high-dimensional vector representations using the bge-m3 embedding model. Milvus' IVF-SQ8 index is employed to perform efficient similarity searches on these vectors, calculating the semantic similarity between them and the generated text. In this way, the references most relevant to the generated content can be quickly located, and a similarity score can be obtained. To process the original references, sentences are first segmented based on regular punctuation marks. The processed sentences are then grouped into three sets, with each set containing an overlapping sentence. This overlapping method helps retain richer information, improving recall integrity, and ensures the final block has semantic coherence. To effectively manage large-scale vector similarity searches, we integrated Milvus, a high-performance open-source vector database optimized for handling large volumes of unstructured data. Milvus supports storage, indexing, and fast retrieval of document block embeddings, ensuring that similarity calculations remain fast and scalable.

\paragraph{Head and Tail Generation} 
Following the generation of section content, the head and tail generation module is responsible for creating the introduction and conclusion parts of the book. This module uses the generated section content as reference material and employs Large Language Models (LLMs) to craft the introductory and concluding sections of the book, providing a cohesive narrative that frames the entire work.

\subsection{System Walkthrough}
Figure~\ref{fig:demonstration} gives a comprehensive walkthrough of how the system can be operated by users.
\begin{itemize}
    \item \textbf{Step 1:} Users upload the outline, specify sections, and upload relevant reference documents.;
    \item \textbf{Step 2:} Use the PDF parsing tool to extract content from PDF documents;
    \item \textbf{Step 3:} Use LLMs to compress the content of the parsed documents.
    \item \textbf{Step 4:} Use LLMs to generate content for sections, and experts can manually edit the generated content in the editing area.
    \item \textbf{Step 5:} Conduct reference verification for the generated content to ensure the accuracy of citations.
\end{itemize}

\section{Evaluation}
In terms of evaluation, we adopt a phased approach to evaluate the system. First, we use the system to perform literature review tasks, assessing the model’s ability to produce structured outputs and the readability of the generated content. Once the system’s usability is confirmed, we further evaluate its capability in book-writing tasks during the human–AI collaborative generation of \textit{AI for Rock Dynamics}.

\subsection{Literature Review Evaluation}
\paragraph{Datasets} We have collected 500 English scientific research reviews\footnote{https://github.com/Kexin-Technology/EnSciRL-500}. Table \ref{tab:appendix_example_1} provides a data example from one of the scientific literature reviews. In addition, we have further extracted the outlines of the references to evaluate the outlines generated by the large language models subsequently.

\paragraph{Implementation Setups} During the experiments, we selected several mainstream large language models from both domestic and international sources. We observed that, regardless of prompt adjustments, LLMs tend to produce subsection-style outputs (with subheadings) when generating long-form text. Therefore, we focused on evaluating LLMs’ ability to generate outlines in the literature review task. First, we directly fed metadata (title, subject, references) into LLMs to generate an initial outline of the literature review. Then, following the \emph{section generation} procedure, we generated the content for each section, ensuring that the output was both relevant and supported by existing literature. Finally, we stitched together the generated text from each section to form a complete and coherent literature review, which was then used for evaluating the generated content.

\paragraph{Automatic Evaluation Results} We selected the \emph{ROUGE-1/2/L} provided by Google\footnote{https://github.com/google-research/google-research/tree/master/rouge} to evaluate the content and the \emph{Soft Heading Recall} (S-H Recall) \cite{FRANTI2023115} for evaluating the outline of generated survey.

\begin{equation}
\begin{split}
\text{Sim} \left( t_i, t_j \right) = \cos \left( \text{embed} \left( t_i \right), \text{embed} \left( t_j \right) \right)
\end{split}
\end{equation}

\begin{equation}
\begin{split}
\text{card}(T) = \sum_{i=1}^{|T|} \frac{1}{\sum_{j=1}^{|T|} \text{Sim} \left( t_i, t_j \right)}
\end{split}
\end{equation}

\begin{equation}
\begin{split}
\text{card}(R \cap G) = \text{card}(R) + \text{card}(G) - \text{card}(R \cup G)
\end{split}
\end{equation}

\begin{equation}
\begin{split}
\text{soft heading recall} = \frac{\text{card}(R \cap G)}{\text{card}(R)}
\end{split}
\end{equation}
$T = \{ t_1, t_2, t_3, \cdots, t_K \}$ represents a group of the chapter titles/heads in a generated/reference survey. R and G are the chapter titles of the generated and reference survey, respectively. The bge-large-en-v1.5\footnote{https://huggingface.co/BAAI/bge-large-en-v1.5} model is used for text embedding. This score encourages the similarity between generated and reference chapter titles while punishes the similarity of titles within the generated survey.

\begin{table*}[ht]
    \centering
     \begin{tabular}{c|c|c|c|c}
    \hline
         LLMs used 
& S-H Recall& ROUGE-1& ROUGE-2& ROUGE-L\\
    \hline
    claude-3.5-sonnet& 0.9802&	0.4932&	0.1504& 0.1467\\
    GPT-4o& 0.9721&	0.4603&	0.1417& 0.1416\\
    KIMI& 0.9604&	0.4423&	0.1459& 0.1301\\
    GPT-4-turbo& 0.9576&	0.4136&	0.1357& 0.1351\\
    Qwen2.5-70B-Instruct& 0.9501& 0.3903&	0.1274& 0.1293\\
    GPT-3.5-turbo& 0.8843&	0.2651&	0.0783& 0.0983\\
    GLM4-9B& 0.8309&	0.2713&	0.0613& 0.0994\\
    \hline
    \end{tabular}
    \caption{Performance Comparison of Large Language Models on S-H Recall and ROUGE Metrics}
    \label{tab:final result}
\end{table*}

Table \ref{tab:final result} presents the scores of different models in the task of literature review generation. The evaluation metrics include the ROUGE score and the Soft Heading Recall (S-H Recall) score. It can be observed that the Claude model performs better in both S-H Recall and ROUGE scores, achieving an especially high S-H Recall score of 0.9802. It leads across all evaluation metrics and secures the top position. These automatic evaluation metrics indicate that the system is desirable in terms of outline generation capability and text coherence.

\paragraph{Human Evaluation Results} cover five aspects: (i) Fluent and clear language; (ii) Logical structure; (iii) reliable citations; (iv) Consistency of content with the theme; (v) Broad analytical scope. 

\begin{table*}[ht]
    \centering
     \begin{tabular}{c|c|c|c|c|c} \hline  
         LLMs used 
&  Prim-eval&  Sec-eval&    Fir-exam&Fin-exam&Avg\_score\\ \hline  
         claude-3.5-sonnet&  0.853&  0.789&    0.691&A&0.821 
\\ 
         GPT-4o&  0.788&  0.673&    0.621&B&0.731 
\\ 
         KIMI&  0.751&  0.636&    0.632&B&0.694 
\\ 
         GPT-4-turbo&  0.669&  0.623&    0.621&C&0.646 
\\ 
         Qwen2.5-70B-Instruct&  0.616&  0.587&    0.613&C&0.602 
\\  
         GPT-3.5-turbo&  0.603&  0.511&    0.542&D&0.557 
\\ 
         GLM4-9B&  0.541&  0.501&    0.368&E&0.521 
\\ 
         \hline
    \end{tabular}
    \caption{Results of Human Evaluation}
    \label{tab:human_score}
\end{table*}

Table \ref{tab:human_score} presents the results of the human evaluation of 20 articles by a 5-member team. It was followed a structured pipeline: one primary evaluator (Prim-eval) scored all articles on the five aspects; two secondary evaluators (Sec-eval) jointly assessed the articles and averaged their scores; one first examiner (Fir-exam) randomly checked 20 articles to ensure quality; and a final examiner (Fin-exam) reviewed the overall results and assigned team performance levels (A–E). The table reveals a high degree of consistency in the human evaluations. After careful deliberation and consensus within the evaluation team, the final human evaluation scores were determined by averaging the scores given by the primary evaluator and the secondary evaluators.

\subsection{Book Writing Evaluation}
Based on the results of the literature review experiments, we validated the usability of \emph{CoAuthorAI} for long-form document generation. The system, in combination with Kexin Technology’s LUFFA model, completed \textit{AI for Rock Dynamics} under the support of the Artificial Pen Project. Since book evaluation is a heavy and complex task, in this evaluation we only present some comparative results between the machine-generated draft and the final draft of \textit{AI for Rock Dynamics}.

\paragraph{Datasets} The book \emph{AI for Rock Dynamics} contains an average of around 130 references per chapter across Chapters 2–8, totaling 910 references. Excluding the Introduction and Conclusion chapters, whose references are the seven generated chapters themselves, the entire book includes 917 references in total.

\paragraph{Implementation Setups} Together with the author team, we iteratively developed the prompts for text compression and section generation. We strictly followed the procedure outlined in Section \emph{3.3 System Walkthrough} to generate each chapter, and used the generated sections as reference material to produce the Introduction and Conclusion chapters.

\paragraph{Evaluation Results} Table~\ref{tab:Citation Accuracy} presents comparative results between the machine-generated draft and the final draft, including citation accuracy—the proportion of traceable citations among all citations produced by the LLM—and the manual correction rate for each chapter conducted by the author team.

To compute the correction rate, both the machine-generated draft and the final draft are segmented at the sentence level, resulting in $n$ sentences in the initial draft and $m$ sentences in the final version. For each corrected chapter, we iterate through the sentences of the final draft and count how many of them also appear in the initial draft; let $s$ denote this count. The correction rate is then defined as:

\begin{equation}
    \text{Correction Rate} = \frac{n - s}{s}.
\end{equation}

\begin{table}[ht]
        \centering
        \small
        \begin{tabular}{c|cc}
        \hline
            Chapter&\ citation accuracy&\ correction rate \\
            \hline
                Chapter 2& 73\%& 16\%\\
                Chapter 3& 76\%& 18\%\\
                Chapter 4& 82\%&21\%\\
                Chapter 5& 81\%&13\%\\
                Chapter 6& 79\%&11\%\\
                Chapter 7& 77\%&15\%\\
                Chapter 8& 74\%&14\%\\
                \textbf{average} & 77.4\%&15.4\%\\
            \hline
        \end{tabular}
        \caption{Citation Accuracy and Manual Correction Rate per Chapter}
        \label{tab:Citation Accuracy}
        \end{table} 

The results show that the average citation accuracy after system verification reaches 77.4\%, mitigating the impact of LLM hallucinations. The manual correction rate fluctuates between 11\% and 21\%, with an average of 15.4\%. These findings indicate that \emph{CoAuthorAI} can achieve satisfactory performance in book writing, though a moderate level of manual intervention is still required to ensure reliability.

\section{Conclusion}
The \emph{CoAuthorAI} system offers a novel and effective approach to scientific book writing by integrating human expertise with the capabilities of large language models. By leveraging human guidance through expert-crafted outlines and iterative feedback loops, the system ensures the quality and precision of the generated content. The CoAuthorAI system has the potential to significantly streamline the scientific writing process and contribute to the production of high-quality scientific books. Future work may focus on further refining the system's capabilities and exploring its applications in other scientific writing tasks.

\section*{Limitations}
Despite continuous adjustments to the prompts in Table \ref{tab:appendix_example_2}, \emph{CoAuthorAI} still retains a typical machine-generated format in the final books. The lack of visual elements such as images and tables reduces the readability and appeal of the books. Compared with traditional books, the content generated from existing literature lacks innovative materials and mainly consists of summaries and syntheses of past knowledge.


\bibliography{custom}

\appendix
\section{Appendix}

\begin{table*}[ht]
    \centering
    \begin{tabular}{c|p{13cm}}
    \hline
    Key & Values  \\
    \hline
        title &A formula for a quartic integral: a survey of old proofs and some new ones 
        \\ 
        article\_id & arXiv:0707.2118
        \\ 
        subject &["Classical Analysis and ODEs"]
        \\
        abstract&We discuss several existing proofs of the value of a quartic integral and present a new proof that evolved from rational Landen ...
        \\
       content &1. Introduction The evaluation of deﬁnite integrals has attracted the scientiﬁc community, both professional and amateurs, for a long ...
        \\
       reference &[
       [1] B. Berndt. Ramanujan’s Notebooks, Part I. Springer-Verlag, New York, 1985.,

       [2] G. Boros and V. Moll. An integral hidden in Gradshteyn and Ryzhik. Jour. Comp. Applied Math., 106:361–368, 1999.,
       ...
       ]\\
       reference\_content &[\{
       reference\_num: [2],
       reference\_title: An integral hidden in Gradshteyn and Ryzhik,
       reference\_abstract: We provide a closed-form expression for the integral ...
       \},
       ...]
        \\
        \hline 
\end{tabular}
\caption{An Example from datasets }
\label{tab:appendix_example_1}
\end{table*}

\begin{table*}[ht]
    \centering
    \begin{tabular}{c|p{13cm}}
    \hline
    Prompts & Content  \\
    \hline
        Compression Prompt &  Write a detailed research report in English aimed at helping users quickly understand and grasp the core content and key findings of the provided paper.

When writing the research report, please follow these steps:

1. Initial Understanding: First, gain a preliminary understanding of the research area or direction provided by the user to ensure clarity regarding their needs.

2. Paper Analysis: Carefully read and analyze the full text of the paper to identify all key research assumptions, princinples, formulas,methods, results, and conclusions.

3. Key Points Extraction: Extract the most important points from the paper, including but not limited to experimental design, data analysis, main findings, conclusions, and formulas.

4. Language Expression: Use precise, clear, and professional language, avoiding vague or ambiguous expressions.

5. Proofreading and Revision: After completing the draft, carefully proofread and revise to ensure there are no grammatical or spelling errors, while also ensuring the accuracy and completeness of the report.

The specific requirements for the research report are as follows (!!! Each point is very important !!!):

* The length of the research report should be around 4,000 words, with substantial content, avoiding hollow or meaningless sentences.

* The report should cover all important points from the paper, ensuring comprehensiveness and avoiding excessive detail on any one specific point while omitting other important content.

* The report should retain the most important formulas, tables, and key data from the paper so that users can fully understand the research results. Use markdown format to write these important formulas and tables in the report!!!

* Remember, users are interested in the technical points and important findings in the required research paper; the research report should not include content such as authors, acknowledgments, or future directions!!!

        \\ 
        Generation Prompt& You are a distinguished scholar specializing in writing comprehensive review books. Your current task is to assist users in crafting a specific section of their review book with meticulous attention to detail and scholarly rigor.

Input:
The user will provide you with:
1. The overall topic, outline, and title of their review book
2. The hierarchical headings for the section you're tasked to write
3. Titles and abstracts of reference papers (in English or Chinese) to be used for this section

Your focus:
Write the content for the lowest-level section heading provided (e.g., if given headings up to the third level, you'll write content for the third-level heading).
Requirements:

1. Language: Write in English, regardless of the language of the references.

2. Citations: Use the format "[idx]" or "[idx\_1, idx\_2, ...]" (e.g., [3] or [3, 49]).

3. Content: 
   - Focus solely on the body text; exclude section headings, reference lists, or other supplementary content.
   
   - Write comprehensively, aiming for a minimum of 8000 words.
   
   - Describe all relevant details extensively.
   
   - Include important formulas and tables from the references.
   
   - Formulas can be screen on the makrdown file
   
4. References: Ensure each provided reference is cited at least once.

5. Style: Write as part of a larger work, avoiding introductory or concluding sentences that might be more appropriate for a standalone piece.
\\
\hline 
\end{tabular}
\caption{Prompts for the CoAuthorAI}
\label{tab:appendix_example_2}
\end{table*}

\end{document}